\documentclass[wcp]{jmlr}

\usepackage{longtable}

\usepackage{booktabs}
\usepackage{multirow}

\jmlrvolume{80}
\jmlryear{2017}
\jmlrworkshop{ACML 2017}

\title[PPMN]{Pyramid Person Matching Network for Person Re-identification}

  \author{\Name{Chaojie Mao} \Email{mcj@zju.edu.cn}\\
  \Name{Yingming Li} \Email{yingming@zju.edu.cn}\\
  \Name{Zhongfei Zhang} \Email{zhongfei@zju.edu.cn}\\
  \Name{Yaqing Zhang} \Email{yaqing@zju.edu.cn}\\
  \addr College of Information Science and Electronic Engineering, Zhejiang University, Hangzhou, China \\
  \Name{Xi Li} \Email{xilizju@zju.edu.cn}\\
  \addr College of Computer Science and Technology, Zhejiang University, Hangzhou, China \\
}

\begin{document}

\maketitle

\begin{abstract}
In this work, we present a deep convolutional pyramid person matching network (PPMN) with specially designed Pyramid Matching Module to address the problem of person re-identification. The architecture takes a pair of RGB images as input, and outputs a similiarity value indicating whether the two input images represent the same person or not. Based on deep convolutional neural networks, our approach first learns the discriminative semantic representation with the semantic-component-aware features for persons and then employs the Pyramid Matching Module to match the common semantic-components of persons, which is robust to the variation of spatial scales and misalignment of locations posed by viewpoint changes. The above two processes are jointly optimized via a unified end-to-end deep learning scheme. Extensive experiments on several benchmark datasets demonstrate the effectiveness of our approach against the state-of-the-art approaches, especially on the rank-1 recognition rate.
\end{abstract}
\begin{keywords}
Person re-identification, Pyramid Matching Module, Unified end-to-end deep learning scheme
\end{keywords}

\section{Introduction}

The task of person re-identification (Re-ID) is to judge whether two person images represent the same person or not and it has widely-spread applications in video surveillance. There are two challenges posed by viewpoint changes: the variation of a person's pose and misalignment. 

\begin{figure}
\includegraphics[width=1\textwidth]{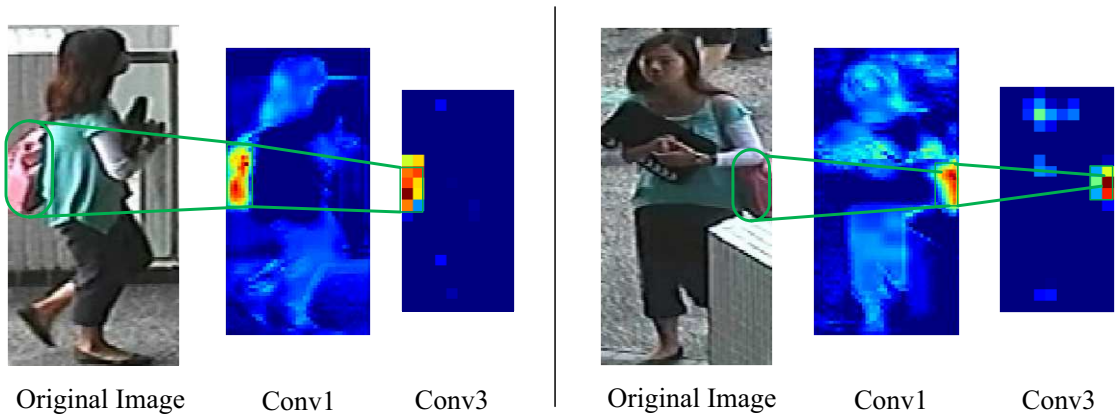}
\caption{An example of misalignment for the component ``bag'' in two different images where the feature maps are extracted with the GoogLeNet.}
\label{fig:example}
\end{figure}

Many existing methods solve the challenges above by extracting cross-view invariant features~\cite{Alpher13}~\cite{Alpher14}~\cite{Alpher16}~\cite{Alpher27}~\cite{Alpher19}. These methods focus on extracting local features including the hand-crafted features and deep learning features from horizontal stripes of a person image, and fuse them into a description vector as the representation. Though these methods usually work under the assumption up to a slight vertical misalignment, they ignore the  typically widely existing horizonal misalignment. From the perception of humans, the images captured by two cameras for the same person should have many common components (body-part, front and back pose, belongings) so that people can decide whether the two input images represent the same person or not. Based on this principle, methods like DCSL~\cite{Alpher06} employ deep convolutional networks to learn the correspondence among these components and have shown a promising performance. DCSL uses the deep convolutional networks like GoogLeNet~\cite{Alpher09} to extract the semantic-components representation. For bottom layers, the discriminative region in each feature map learned by DCSL corresponds to one component of a person such as bag, head, and body. For high layers, the learned regions still keep their shapes and spatial locations while they are abstract. However, the feature regions of the same components from two views for the same person seldom have the consistent spatial scales and locations because of viewpoint changes. For example, the component ``bag'' is located in the opposite sides in the two images in Figure~\ref{fig:example}. Consequentely, the existing methods like DCSL ignore this problem.  

To address the challenges above, in this work, we present a deep convolutional pyramid person matching network (PPMN). Pyramid matching based on convolution operation is employed to compute the responses of the same semantic-component in different images. To further capture the variation of the spatial scale and location misalignment, we exploit the flexible kernel-size convolution operation to guarantee that the most of semantic-components are matched in the same subwindows. Since the convolution operation with large kernels increases the parameters and computation, we propose to reduce the computation complexity by introducing the atrous convolution structure~\cite{Alpher23}, which has been used in some convNet-based tasks such as image segmentation, object detection, and can provide a desirable view of perception without increasing parameters and computation by introducing zeros between the consecutive filter values. In particular, we employ the multi-rate atrous convolution layers to construct the Pyramid Matching Module and produce the correspondence representation between the semantic-components. With the correspondence representation, we learn the final similiarity value to decide whether the two input images represent the same person or not.  

The proposed framework is evaluated on three real-world datasets. Extensive experiments on these benchmark datasets demonstrate the effectiveness of our approach against the state-of-the-art, especially on the rank-1 recognition rate.

The main contributions of this paper are as follows:
(1) We propose an end-to-end deep convolutional framework to deal with the problem of person Re-ID. Image representation learning and cross-person correspondence learning are jointly optimized to enable the image representation to adapt to the task of perosn Re-ID.
(2) The proposed framework maps a person's semantic-components to the deep feature space and employs the pyrimid matching strategy based on the atrous convolution to identify the common components of the person.

\section{Related Work}
In the literature, most existing efforts of person Re-ID are mainly carried in two aspects: the discriminative representation learning and the effective matching strategy learning. For image representation, a number of approaches pay attention to designing robust descriptors againist misalignments and variations. Early studies employ hand-crafted features including HSV color histogram~\cite{Alpher01}, SIFT~\cite{Alpher02}, LBP~\cite{Alpher03} features or the combination of them.  Recently, several deep convolutional architectures \cite{Alpher27}~\cite{Alpher06} have been proposed for person Re-ID and have shown significant improvements over those with hand-crafted features. 

For matching strategy, the essential idea behind metric learning is to find a mapping function from the feature space to the distance space so as to minimize the intra-personal variance while maximizing the inter-personal margin. Many approaches have been proposed based on this idea including LMNN~\cite{Alpher13} and KISSME~\cite{Alpher14}. Recently, some efforts jointly learn the representation and classifier in a unified deep architecture. For example, patch-based methods~\cite{Alpher16}~\cite{Alpher27} decompose images into patches and perform patchwise distance measurement to find the spatial relationship. Part-based methods~\cite{Alpher19} divide one person into equal parts and jointly perform bodywise and partwise correspondence learning since the pedestrians keep upright in general. Different from all the above efforts which focus on feature distance measurement, our proposed method aims at learning the semantic correspondence of semantic-components based on the semantics-aware features and is robust to the variation and misalignment posed by viewpoint changes. 

\section{Our Architecture}
Figure \ref{fig:architecture} illustrates our network's architecture. The proposed architecture extracts the semantics-aware representations for a pair of input person images. The features are then concatenated to feed into the Pyramid Matching Module to learn the correspondence of semantic-components. Finally, softmax activations are employed to compute the final decision which indicates the probability that the image pair represents the same person. The details of the architecture are explained in the following subsections.   
\subsection{Learning representation for Images}
\begin{figure}
\includegraphics[width=1\textwidth]{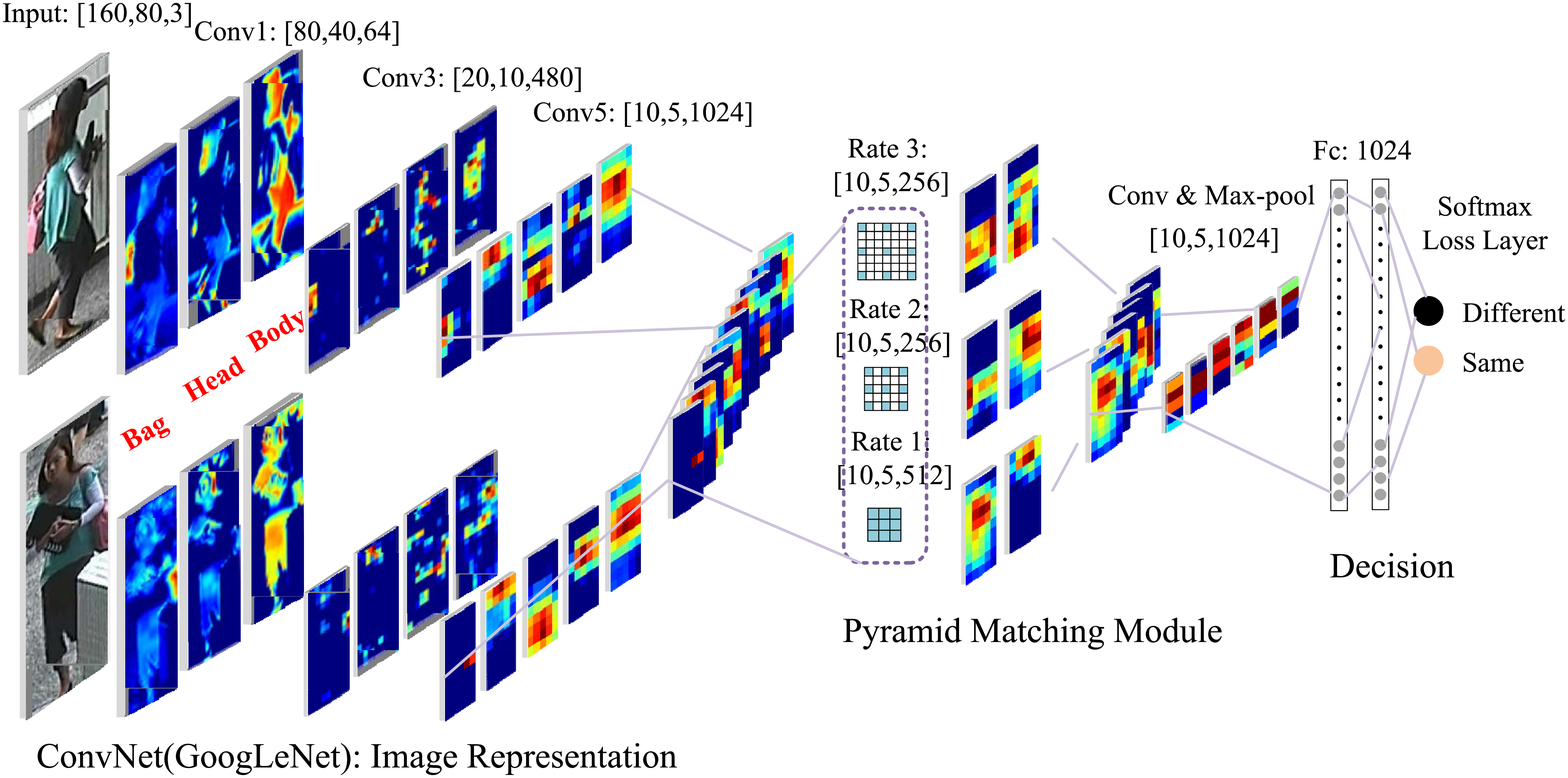}
\caption{The proposed architecture of the deep convolutional Pyramid Person Matching Network (PPMN). Given a pair of person images as input, the parameters-shared GoogLeNets generate the semantics-aware representation. The semantic components such as bag and head are visible in the output of Conv1 layer.  With the extracted features, the Pyramid Matching Module learns the correspondence of these semantic components based on multi-scale artrous convolution layers. Finally, softmax activations give the final decision of whether the image pair depicts the same person or not.}
\label{fig:architecture}
\vspace*{-2em}
\end{figure}

The ImageNet-pretrained GoogLeNet employed in this work is able to capture the semantic features for most of objects in this task as the ImageNet dataset contains a large number of object types for more than 100000 concepts. In our architecture, these semantic features are extracted with two parameter-shared GoogLeNets for a pair person images, respectively. As shown in Figure \ref{fig:architecture}, the GoogLeNets have been adapted to the Re-ID task by finetuning on a Re-ID dataset and decompose the person image into many semantic components such as bag, head, and body. It is apparent to recognize the particular components from the visualization of bottom layers' output like Conv1 layer. These components' visualizations for higher layers such as Conv5 layer are more abstract but still keep the shapes and spatial locations. For notational simplicity, we refer to the convNet as a function $f_{CNN}( \boldsymbol X;  \boldsymbol \theta)$, which takes $ \boldsymbol X$ as input and $ \boldsymbol \theta$ as parameters. Given an input pair of images resized to $160\times80$ from two cameras, A and B, the GoogLeNets output 1024 feature maps with size $10\times5$ separately as the representations of images. We denote this process as follows:
\begin{equation}
\label{equ:imgPre}
\{\boldsymbol R^A, \boldsymbol R^B\}=\{f_{CNN}(\boldsymbol I^A; \boldsymbol \theta_1), f_{CNN}(\boldsymbol I^B; \boldsymbol \theta_1)\}
\end{equation}
where $\boldsymbol R^A$ and $\boldsymbol R^B$ denote the representations of images $\boldsymbol I^A$ and $\boldsymbol I^B$, respectively. $\boldsymbol \theta_1$ are the shared parameters. 

\subsection{Pyramid Matching Module using Atrous Convolution}
\begin{figure}
\includegraphics[width=1\textwidth]{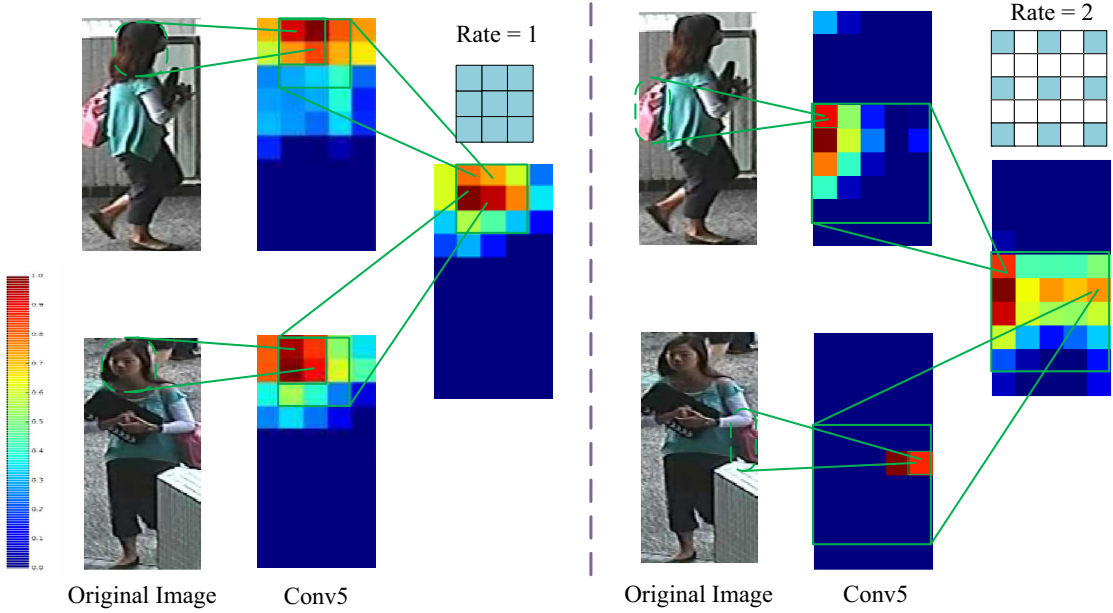}
\caption{
Illustration of the correspondence learning with Pyramid Matching Module. Left: the component ``head'' has the similiar spatial shapes and locations. Right: the component ``bag'' has the completely different shapes and locations. We match the components above by computing their responses in the corresponding windows and take the convolutions with a multi-scale field-of-view, which are robust to the misalignments of locations and variations of scale posed by viewpoint changes.
}
\label{fig:fusion_layer}
\end{figure}

Based on the semantic representations of persons, the problem of person matching is reduced to a problem of the matching for the semantic-components. However, the challenges are the variations of spatial scales and the misalignments of locations for the semantic-components posed by viewpoint changes. As shown in Figure \ref{fig:example}, the same bag belonging to the same person is located on the right side in one image but the left side in the other image. To deal with the challenges, we employ the atrous convolution with multi-scale kernels to construct a module called Pyramid Matching Module based on the pyramid matching strategy. We give two examples in Figure~\ref{fig:fusion_layer} to explain how this module works. On the left column, the component ``head'' has the similiar spatial shapes and locations in the two images. It is easy to learn the correspondence between the two feature maps with a general convolution operation to compute the responses of two feature regions in closely located windows in the two images, respectively, called the field-of-views, while the component ``bag'' has completely different shapes and locations in the two images, respectively, and thus has different field-of-views. Consequently, a larger field-of-view is required for the convolution in the latter case. Accordingly, we employ the atrous convolution for a large field-of-view. The Pyramid Matching Module includes three branches $3\times3$ atrous convolution with rate 1, 2 and 3, respectively, which provides the field-of-view with size $3\times3$, $5\times5$, $7\times7$, respectively. With the images concatenated as $\{\boldsymbol R^A,\boldsymbol R^B\}$, the proposed module computes the correspondence distribution denoted as $\boldsymbol S_{PPM} = \{ \boldsymbol S_{r=1}$, $\boldsymbol S_{r=2}$, $\boldsymbol S_{r=3} \}$, in which the value of each location $(i, j)$ indicates the correspondence probability at that location. $r$ is the rate of atrous convolution. We formulate this matching strategy as follows:
\begin{align}
\label{equ:ppm}
 \boldsymbol S_{PPM} ={} & \{\boldsymbol S_{r=1}, \boldsymbol S_{r=2}, \boldsymbol S_{r=3}\} \notag \\
={} & \{f_{CNN}(\{\boldsymbol R^A,\boldsymbol R^B\}; \{\boldsymbol \theta^1_2, \boldsymbol \theta^2_2, \boldsymbol \theta^3_2\}\} 
\end{align}
where $\boldsymbol \theta^r_2(r=1,2,3)$ are the parameters of the matching branch with rate $r$. We use $\boldsymbol \theta_2 = \{ \boldsymbol \theta^1_2, \boldsymbol \theta^2_2, \boldsymbol \theta^3_2 \}$ as the parameters of our module.

We fuse the concatenated correspondence maps $S_{PPM}$ with the learned parameters $\boldsymbol \theta_3$, which indicates the weights of different matching branches, and output the fused correspondence representation $\boldsymbol S_{fusion}$. Inspired by \cite{Alpher06}, we further downsample $S_{fusion}$ by the max-pooling operation so as to preserve the most discriminative correspondence information and align the result in a larger region. Then, we obtain the final correspondence representation $\boldsymbol S_{final}$:
\begin{align}
\label{equ:weights}
\boldsymbol S_{final} = f_{CNN}(\{\boldsymbol S_{r=1}, \boldsymbol S_{r=2}, \boldsymbol S_{r=3}\}; \boldsymbol \theta_3) 
\end{align}


\subsection{The unified framework and Learning}

We apply two fully connected layers to encode the correspondence representation $\boldsymbol S_{final}$ with an abstract vector of size 1024. The vector is then passed to a softmax layer with two softmax units $\boldsymbol S(\boldsymbol S_{final}; \boldsymbol \theta_4)$: namely $\boldsymbol S_0(\boldsymbol S_{final}; \boldsymbol \theta_4)$ and $\boldsymbol S_1(\boldsymbol S_{final}; \boldsymbol \theta_4)$. We represent the probability that the two images in the pair, $\boldsymbol I^A$ and $\boldsymbol I^B$, are of the same person with softmax activations computed on the units above:
\begin{equation}
\label{equ:softmax}
p = \frac {exp(\boldsymbol S_1(\boldsymbol S_{final}; \boldsymbol \theta_4))}{exp(\boldsymbol S_0(\boldsymbol S_{final}; \boldsymbol \theta_4))+exp(\boldsymbol S_1(\boldsymbol S_{final}; \boldsymbol \theta_4))}
\end{equation}

We reformulate this approach as a unified framework with $\boldsymbol \theta = \{\boldsymbol \theta_1, \{\boldsymbol \theta^r_2\}, \boldsymbol \theta_3, \boldsymbol \theta_4 \}$, where $r=1,2,3$ based on Eqs.\ref{equ:imgPre} - \ref{equ:weights} :    
\begin{align}
\label{equ:unified} 
 S(\boldsymbol S_{final}, \boldsymbol \theta_4) ={} & f_{CNN}(\{\boldsymbol S_{r=1}, \boldsymbol S_{r=2}, \boldsymbol S_{r=3}\}; \boldsymbol \theta_4, \boldsymbol \theta_3) \notag \\
={} & f_{CNN}(\{\boldsymbol I^A,\boldsymbol I^B\}; \boldsymbol \theta_4, \boldsymbol \theta_3, \{\boldsymbol \theta^r_2\},\boldsymbol \theta_1) \notag \\
={} & f_{CNN}(\{\boldsymbol I^A,\boldsymbol I^B\}; \boldsymbol \theta)
\end{align}
We optimize this framework by minimizing the widely used cross-entropy loss over a training set of $N$ pairs:
\begin{align}
\label{equ:loss} 
\boldsymbol L_\theta = - \frac{1}{N} \sum^N_{n=1} [ l_n \log p_n + (1-l_n) \log (1-p_n) ]
\end{align}
where $l_n$ is the 1/0 label for the input pair, which represents the same person or not. 

\section{Experiments}
\subsection{Datasets and Protocol}

We compare our proposed architecture with the state-of-the-art approaches on three person Re-ID datasets, namely CUHK03~\cite{Alpher27}, CUHK01~\cite{Alpher28} and VIPeR~\cite{Alpher29}. All the approaches are evaluated with Cumulative Matching Characteristics (CMC) by single-shot results, which characterize a ranking result for every image in the gallery given a probe image. Our experiments are conducted on the datasets with 10 random initializations in training and the average results are provided. Table~\ref{table:dataset} lists the description of each dataset and our experimental settings with the training and testing splits. 

\begin{table}
\caption{Datasets and settings in our experiments. The settings for CUHK01 dataset include the 100 test IDs and 486 test IDs. }
\label{table:dataset}
\begin{center}
\begin{tabular}{c|c|c|c}
	\toprule
	 Dataset & CUHK03 & CUHK01 & VIPeR \\
	\midrule
	 identities & 1360 & 971 & 632 \\
	 images & 13164 & 3884 & 1264 \\
	 views & 2 & 2 & 2 \\
	 train IDs & 1160 & 871;485 & 316 \\
	 test IDs & 100 & 100;486 & 316 \\
	\bottomrule
\end{tabular}
\end{center} 
\end{table}

\subsection{Training the Network}
The proposed architecture is implemented on the widely used deep learning framework Caffe\cite{Alpher30} with an NVIDIA TITAN X GPU. It takes about 40-48 hours in training for 160K iterations with batch size 100. We use stochastic gradient descent for updating the weights of the network. We set the momentum as $\gamma = 0.9$ and set the weight decay as $\mu = 0.0002$. We start with a base learning rate of $\eta^{0} = 0.01$ and gradually decrease it as the training progresses using a polynomial decay policy with $power$ as $0.5$. 
 
\textbf{Data Augmentation}. To make the model robust to the image translation variance and to further augment the training dataset, for every original training image, we sample 5 images around the image center, with translation drawn from a uniform distribution in the range $[-8,8]\times[-4,4]$ for an original image of size $160\times80$.

\textbf{Hard Negative Mining (hnm)}. The negative pairs are far more than the positive pairs, which can lead to data imbalance. Also, in these negative pairs, there still exist scenarios that are hard to distinguish. To address these difficuties, we first sample the negative sets to get three times as many negatives as positives and train our network. Then, we use the trained model to classify all the negative pairs and retain those ranked top on which the trained model performs the worst for retraining the network.

\begin{table}
\begin{center}
  \caption{Comparison of state-of-the-art results on CUHK03. The cumulative matching scores (\%) at rank 1, 5, and 10 are listed.}

  \label{table:CUHK03}
  \begin{tabular}{c|ccc|ccc}
    \toprule
    \multirow{2}*{Methods} & 
    \multicolumn{3}{c|}{labelled CUHK03} &
    \multicolumn{3}{c}{detected CUHK03} \\
    \cline{2-7}
     & r=1 & r=5 & r=10 & r=1 & r=5 & r=10  \\ 
    \midrule
     KISSME & 14.17 & 37.46 & 52.20 & 11.70 & 33.45 & 45.69 \\
     LMNN  & 7.29 & 19.64 & 30.74 & 6.25 & 17.87 & 26.60  \\
     LOMO+LSTM & - & - & - & 57.30 & 80.10 & 88.30  \\
     LOMO+XQDA & 52.20 & 82.23 & 92.14 & 46.25 & 78.90 & 88.55 \\ 
    \midrule
     FPNN & 20.65 & 50.94 & 67.01 & 19.89 & 49.41 & - \\
     ImprovedDL & 54.74 & 86.50 & 93.88 & 44.96 & 76.01 & 81.85 \\
     PIE(R)+Kissme & - & - & - & 67.10 & 92.20 & 96.60 \\
     SICIR & - & - & - & 52.17 & - & - \\
     DCSL(no hnm) & 78.60 & 97.76 & 99.30  \\
     DCSL(hnm) & 80.20 & 97.73 & 99.17 & - & - & - \\ 
    \midrule
     PPMN(no hnm) & 83.20 & 97.50 & 99.25 & 77.60 & \textbf{96.10} & \textbf{98.60} \\
     PPMN(hnm) & \textbf{85.50} & \textbf{98.20} & \textbf{99.50} & \textbf{80.63} & 95.62 & 98.07 \\ 
    \bottomrule
\end{tabular}
\end{center}
\end{table}

\subsection{Experimental Results}
In this section, we campare PPMN with several recent methods, including both hand-crafted features based methods: KISSME~\cite{Alpher14}, LMNN~\cite{Alpher13}, LOMO+LSTM~\cite{Alpher19}, LOMO+XQDA~\cite{Alpher32}; and deep learning features based methods: FPNN~\cite{Alpher27}, ImprovedDL~\cite{Alpher16}, Pose Invariant Embedding (PIE(R)+Kissme)\cite{Alpher34}, Single-Image and Cross-Images Representation learning(SICIR)\cite{Alpher22}, DCSL~\cite{Alpher06}. We report the evaluation results in Table~\ref{table:CUHK03}.

\begin{table}
\begin{center}
  \caption{Comparison of state-of-the-art results on CUHK01 dataset. The cumulative matching scores (\%) at rank 1, 5, and 10 are listed.}
  \label{table:CUHK01}
  \begin{tabular}{c|ccc|ccc}
    \toprule
    \multirow{2}*{Methods} & 
    \multicolumn{3}{c|}{CUHK01(100 test IDs)} &
    \multicolumn{3}{c}{CUHK01(486 test IDs)}  \\ 
    \cline{2-7}
     & r=1 & r=5 & r=10 & r=1 & r=5 & r=10 \\ 
    \midrule
     KISSME & 29.40 & 60.18 & 74.44 & - & - & - \\
     LMNN  & 21.17 & 48.51 & 62.98 & 13.45 & 31.33 & 42.25\\
    \midrule
     FPNN  & 27.87 & 59.64 & 73.53 & - & - & - \\
     ImprovedDL & 65.00 & 89.00 & 94.00 & 47.53 & 71.60 & 80.25 \\
     PIE(R)+Kissme & - & - & - & - & - & - \\
     SICIR & 71.80 & - & - & - & - & - \\
     DCSL(no hnm) & 88.00 & 96.90 & 98.10 & - & - & - \\
     DCSL(hnm) & 89.60 & 97.80 & 98.90 & 76.54 & \textbf{94.24} & 97.49 \\ 
    \midrule
     PPMN(no hnm)  & 92.10 & \textbf{99.50} & \textbf{99.95} & - & - & - \\
     PPMN(hnm) & \textbf{93.10} & 98.80 & 99.80 & \textbf{77.16} & 92.80 & \textbf{97.53} \\ 
    \bottomrule
\end{tabular}
\end{center}
\end{table}

We conduct PPMN on both labelled and detected CUHK03 datasets. From Table~\ref{table:CUHK03}, our method achieves an improvement of 5.30\% (85.50\% vs. 80.20\%) on the labelled dataset and an improvement of 23.33\% (80.63\% vs. 57.30\%) on the detected dataset. Table~\ref{table:CUHK01} also illustrates the top recognition rate on CUHK01 dataset with 100 test IDs and 486 test IDs. We see that PPMN achieves the best rank-1, rank-5 recognition rates of 93.10\%, 99.50\% (vs. 89.60\%, 96.90\% respectively by the next best method) with 100 test IDs, which means in most cases we can find the correct person in the first five samples of the queried and returned results given 100 candidate images. For the settings with 486 test IDs, we finetune the network on the set of half-CUHK01 with the pre-trained model on CUHK03 and achieve an improvement of 0.62\% (77.16\% vs. 76.54\% ) over DCSL using the same training protocol on rank-1 recognition rate. The experimental results also demonstrate the effect of hard negative mining, which provides the absolute gain over 1.00\% compared with the
same model without hard negative mining. Following the setup of~\cite{Alpher16}, we pre-train the network using CUHK03 and CUHK01 datasets, and finetune on the training set of VIPeR. As shown in Table~\ref{table:VIPeR}, we see that PPMN achieves the best rank-1, rank-5, rank-10 recognition rates and an imporvement of 1.20\% (45.82\% vs. 44.62\%) for rank-1 recognition rate. 

\begin{table}
\begin{center}
  \caption{Comparison of state-of-the-art results on VIPeR dataset. The cumulative matching scores (\%) at rank 1, 5, and 10 are listed.}
  \label{table:VIPeR}
  \begin{tabular}{c|ccc}
    \toprule
    \multirow{2}*{Methods} & 
    \multicolumn{3}{c}{VIPeR}\\ 
    \cline{2-4}
     & r=1 & r=5 & r=10 \\ 
    \midrule
     KISSME &19.60 & 48.00 & 62.20\\
     LMNN  & - & - & - \\
     LOMO+LSTM & 42.40 & 68.70 & 79.40 \\
     LOMO+XQDA & 40.00 & 68.13 & 80.51\\ 
    \midrule
     FPNN  & - & - & -\\
     ImprovedDL  & 34.81 & 63.61 & 75.63\\
     PIE(R)+Kissme & 27.44 & 43.01 & 50.82 \\
     SICIR & 35.76 & - & -\\
     DCSL(hnm) & 44.62 & 73.42 & 82.59\\ 
    \midrule
     PPMN(hnm) & \textbf{45.82} & \textbf{74.68} & \textbf{86.08} \\ 
    \bottomrule
\end{tabular}
\end{center}
\end{table}

\section{Conclusion}
In this paper, we have developed a novel deep convolutional architecture for person re-identification. We employ a deep convNet GoogLeNet to map a person's semantic components to the required feature space. Based on the pyramid matching strategy, we design a module to address the misalignment and variation issues posed by viewpoint changes. We demonstrate the effectiveness and promise of our method by reporting extensive evaluations on various datasets. The results have indicated that our method has a remarkable improvement over the state-of-the-art literature.



\bibliography{acml17}

\end{document}